\documentclass[letterpaper, 10 pt, conference]{ieeeconf}  

\IEEEoverridecommandlockouts                              

\usepackage{amsmath,amssymb,mathtools,bm,empheq,gensymb}

\def\bal#1\eal{\begin{align}#1\end{align}} 
\def\suml{\sum\limits}

\newcommand{\etal}{\textit{et al}. }
\newcommand{\ie}{\textit{i}.\textit{e}., }

\newcommand{\br}[1]{\left[#1\right]} 
\newcommand{\pr}[1]{\left(#1\right)} 
\newcommand{\cbr}[1]{\left\{#1\right\}} 
\DeclareMathOperator*{\argmin}{arg\,min} 

\def\transp{\mathsf{T}} 
\def\m{\mathbf}
\def\mc{\mathcal}
\def\R{\mathbb{R}}

\def\ast{*}

\newcommand{\grad}[2]{\ensuremath{\nabla_{#2}#1}} 
\newcommand{\norm}[2]{\ensuremath{\left\|#1\right\|_{#2}}}

\newcommand{\abs}[1]{\ensuremath{\lvert #1\rvert}}
\newcommand {\bbmtx}{\begin{bmatrix}} 
\newcommand {\ebmtx}{\end{bmatrix}} 

\DeclareMathOperator*{\vctr}{vec} 
\newcommand{\vc}[1]{\vctr{\left(#1\right)}}

\DeclareMathOperator*{\diagonal}{diag} 
\newcommand{\diag}[1]{\diagonal\pr{#1}}
\newcommand{\pdiag}[2]{{\diagonal}^{#2}\pr{#1}}

\usepackage[dvipsnames]{xcolor}

\usepackage[ruled]{algorithm2e}
\usepackage{multirow}
\SetKwRepeat{Do}{repeat}{until}
\usepackage{cite}
\usepackage{url}

\title{\LARGE \bf
Robust Two-View Geometry Estimation with Implicit Differentiation}

\author{Vladislav Pyatov$^{1}$, Iaroslav Koshelev$^{2}$ and Stamatios Lefkimmiatis$^{3}$
\thanks{$^{1}$Vladislav Pyatov is with the Skolkovo Institute of Science and Technology (Skoltech), Center for AI Technology, Russia, {\tt\small vladislav.pyatov@skoltech.ru}}.\thanks{$^{2}$
Iaroslav Koshelev is with the AI Foundation and Algorithm Lab, Russia.}%
\thanks{$^{3}$Stamatios Lefkimmiatis is with the MTS AI Group, Russia. {\tt\small s.lefkimmiatis@mts.ai}}
}

\begin{document}

\maketitle
\thispagestyle{empty}
\pagestyle{empty}

\begin{abstract}
We present a novel two-view geometry estimation framework which is based on a differentiable robust loss function fitting. We propose to treat the robust fundamental matrix estimation as an implicit layer, which allows us to avoid backpropagation through time and significantly improves the numerical stability. To take full advantage of the information from the feature matching stage we incorporate learnable weights that depend on the matching confidences. In this way our solution brings together feature extraction, matching and two-view geometry estimation in a unified end-to-end trainable pipeline. We evaluate our approach on the camera pose estimation task in both outdoor and indoor scenarios. The experiments on several datasets show that the proposed method outperforms both classic and learning-based state-of-the-art methods by a large margin. The project webpage is available at: \url{https://github.com/VladPyatov/ihls}
\end{abstract}

\section{INTRODUCTION}
\label{sec:intro}

Nowadays, 3D computer vision has received widespread attention and has found application in a wide range of tasks, including scene reconstruction \cite{Wang_2021_CVPR}, autonomous driving \cite{Hu_2023_CVPR} and neural rendering~\cite{mildenhall2020nerf}. 
To deal with such tasks a  first necessary step is to recover the position and orientation of each camera in the scene or equivalently to solve the Camera Pose Estimation (CPE) problem \cite{Jau_2020_IROS, Parameshwara_2022_CVPR}.

Given a set of input images, the solution of CPE typically consists of three essential steps. First, putative keypoint correspondences are established through the feature extraction \cite{Lowe_2004, Yi_2016, DeTone2017SuperPointSI, Tyszkiewicz_2020_NeurIPS} and matching \cite{Sarlin_2020_CVPR, Sun_2021_CVPR} stages. The second step involves the recovery of the two-view geometry \cite{hartley_zisserman_2004, Szeliski2022} by estimating the relative camera poses between image pairs. This estimate provides a good initial solution that is later refined by bundle adjustment \cite{Triggs1999BundleA} in the final stage. 
The feature matching and bundle adjustment problems have received a lot of attention over the past few years, especially with the advent of learning-based approaches. At the same time two-view geometry estimation has not been widely studied.

The two-view geometry of an image pair is usually represented either with an essential \cite{LonguetHiggins1981ACA} or (in the uncalibrated camera setup) a fundamental matrix \cite{Faugeras1992WhatCB} that define an algebraic representation of the geometric epipolar constraint \cite{hartley_zisserman_2004}. These matrices can be estimated from a set of putative keypoint correspondences using the \textit{least squares} approach. However, due to occlusions, repetitive patterns, illumination changes and poor textures present in a scene, the feature matching stage may produce outliers, mismatched pairs that affect the solution and, above all, lead to an incorrect geometry representation. Given that the least squares method works under the assumption that the noise in the matches obeys a Gaussian distribution, it is unable to efficiently handle severe outliers and can lead to poor estimation results. 

One way to deal with outliers is to utilize RANSAC-based estimators \cite{RANSAC, Torr_2000, Chum2003LocallyOR, Chum2005TwoviewGE, Barath_2019_CVPR, Barath_2020_CVPR}. The main strategy followed by these methods is based on the hypothesize-and-verify approach. Specifically, after the feature matching stage, minimal subsets of the putative matches are randomly sampled and the corresponding two-view geometries are estimated with either the 8-\cite{LonguetHiggins1981ACA} 7- \cite{Hartley_1994_TPAMI} or 5-point \cite{Nister_2004_TPAMI} algorithms in order to form a pool of model hypotheses. Then the generated hypotheses are scored by counting the number of inliers, \ie matches with residual error below a selected threshold. Finally, the best model hypothesis is further refined by using all the inlier correspondences. Despite their effectiveness, the non-deterministic nature of RANSAC-based estimators precludes efficient end-to-end training when neural networks are involved in the overall pipeline. As a consequence, taking full advantage of the information available from the feature extraction and matching stages becomes challenging.

Another line of work that aims to circumvent the problems appearing due to outliers, is devoted to the introduction of robust loss functions \cite{Torr1997TheDA, ruckstuhl2014}, a specific type of M-estimators \cite{Huber1964RobustEO}. According to this approach, the solution to the two-view geometry problem is obtained by optimizing an objective function that is less sensitive to the presence of outliers. In general, the task is formulated as a non-convex and possibly non-smooth optimization problem that is typically solved employing iterative numerical algorithms \cite{Peng_2023_CVPR}. In general, given the non-convex nature of the objective, the solution is not guaranteed to attain the global minimum, but in practice if the fraction of outliers is relatively small, then the recovered two-view geometry is more precise compared to the one estimated with RANSAC-based methods. On the other hand, when the outlier pairs start dominating the correspondences, the solution becomes sub-optimal and hurts the performance of the downstream tasks. 

The learning-based approach in~\cite{Ranftl_2018_ECCV} tries to alleviate this problem by aiming to implicitly learn a robust loss function. However, this method faces the important limitation of having to back-propagate through unrolled layers that involve the singular value decomposition (SVD). This leads both to an increased memory footprint during training as well as numerical instabilities mostly from the back-propagation through SVD. Therefore, the authors of~\cite{Ranftl_2018_ECCV} limit the number of the employed unrolled layers to just a few, which in turn hinders the final performance of their approach. 

In this work we revisit the direction of robust loss function fitting and propose a novel approach for outlier-robust two-view geometry estimation problem. First, we adopt a specific parametric form for the robust loss that exhibits several important benefits both in terms of robustness to outliers and efficiency in training. 
To successfully minimize this non-convex loss we introduce the Iterative Homogeneous Least Squares (IHLS) solver that takes as input putative keypoint correspondences and outputs an estimate of the underlying two-view geometry. To avoid back-propagation through IHLS with unrolling, we take advantage of the convergence guaranties of our solver and utilize the implicit function theorem that allows us to perform back-propagation without the need to save any intermediate results from the forward pass or back-propagate through the SVD layer, which is an essential part of our optimization algorithm. Further, to downplay the role of the outliers, we develop a learnable framework on top of the IHLS solver and utilize a weight prediction network that re-estimates matching confidences in a recurrent fashion based on the matches, residuals of the current solution and auxiliary information from the feature matching stage. We then chain our framework with the LoFTR \cite{Sun_2021_CVPR} dense feature matching transformer to obtain a fully differentiable robust camera pose estimation pipeline. To summarize, our main contributions are as follows:
\begin{itemize}
    \item We employ a robust parametric loss function to deal with the fundamental matrix estimation problem and propose the Iterative Homogeneous Least Squares solver which can efficiently recover a solution of the underlying minimization problem. 
    \item The convergence guaranties of IHLS further allows us to introduce an efficient end-to-end training strategy using implicit backpropagation, which significantly reduces the memory requirements and improves the stability of training compared to alternative approaches that employ unrolled networks.
    \item We propose an end-to-end trainable camera pose estimation pipeline that brings together feature extraction, matching and robust two-view geometry estimation.
\end{itemize}
\begin{figure*}[ht]
    \centering
    \includegraphics[width=\linewidth]{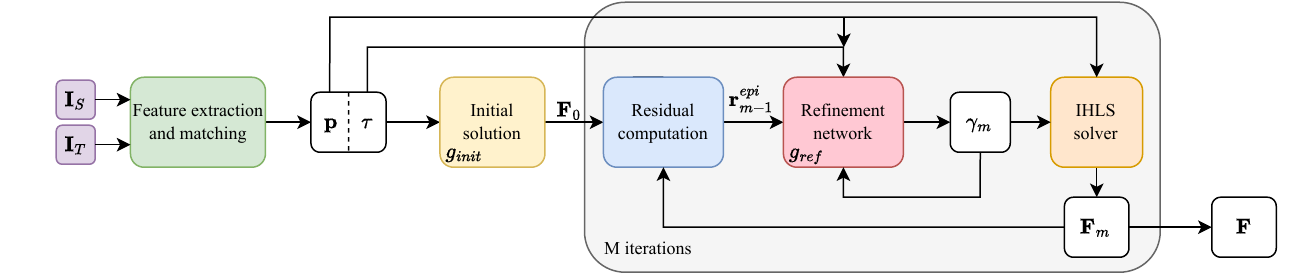}
    \caption{Overview of the proposed framework. The pair of source and target images $\bm I_S, \bm I_T$ is passed through the LoFTR feature extraction and matching module to predict the set of correspondences $\bm p$ and a context vectors $\bm \tau$. Both are then used to initialize and guide the fundamental matrix estimation via the proposed recurrent strategy consisting of refinement network and IHLS solver. For the detailed description of each component please refer to Sec.~\ref{sec:method}.}
    \label{fig:framework}
    \vspace{-0.5cm}
\end{figure*}

\section{RELATED WORK}
\label{sec:related_work}

\subsection{Feature extraction and matching}
Hand-crafted local features, such as SIFT \cite{Lowe_2004} and ORB \cite{Rublee_2011}, combined with a mutual nearest neighbour search or Lowe's ratio test \cite{Lowe_2004} have been considered as the dominant correspondence estimation approach for a long time. Nevertheless, with the advent of deep learning the components of the classic pipeline have been largely revisited. In particular, learning-based local features \cite{Yi_2016, DeTone2017SuperPointSI} show impressive results in challenging conditions of viewpoint change and poor texture. SuperGlue \cite{Sarlin_2020_CVPR} proposes a learning-based approach for local feature matching on top of the sparse SuperPoint \cite{DeTone2017SuperPointSI} features, while the recently introduced LightGlue \cite{lindenberger2023lightglue} further reduces its computational complexity. LoFTR \cite{Sun_2021_CVPR} and its successors \cite{Yu_2023_CVPR} take advantage of dense feature extraction and propose a detector-free design.

\subsection{RANSAC-based estimators}
RANSAC-based approaches either try to improve hypothesis scoring of the original RANSAC \cite{RANSAC} or propose end-to-end trainable counterparts. MLESAC \cite{Torr_2000} chooses the solution that maximizes the likelihood rather than just the number of inlier pairs and thus is less sensitive to the inlier threshold. LO-RANSAC \cite{Chum2003LocallyOR} applies local optimization to the so-far-the-best model hypothesis. DEGENSAC \cite{Chum2005TwoviewGE} detects degenerate configurations and rejects models that are incompatible with the scene geometry. MAGSAC \cite{Barath_2019_CVPR} proposes to marginalize the quality function over a range of noise levels to eliminate the threshold from the model quality calculation.  State-of-the-art approach MAGSAC++ \cite{Barath_2020_CVPR} proposes a novel marginalization procedure and introduces a new model quality function that does not require the inlier-outlier decision. On the other hand, inspired by the success of reinforcement learning, DSAC \cite{Brachmann_2017_CVPR} methods introduce a first learning-based alternative and incorporate REINFORCE \cite{Williams_1992} algorithm to make the hypothesis selection procedure differentiable. Neural-guided RANSAC \cite{Brachmann_2019_ICCV} makes a step forward and proposes learned guidance of hypothesis sampling. $\nabla$-RANSAC \cite{Wei_2023_ICCV} revises the learning-based problem formulation and incorporates the Gumbel Softmax \cite{jang2017categorical} relaxation to estimate the gradients in the sampling distribution.

\subsection{Robust loss functions} 
Robust loss functions serve as an alternative strategy for outlier-robust model fitting and is the main focus of this work. Since their introduction in the context of M-estimators \cite{Huber1964RobustEO}, their applicability to the two-view geometry estimation \cite{MacTavish_2015_AtAC} has been proven. The aforementioned MAGSAC++ \cite{Barath_2020_CVPR} reformulates the model fitting problem as an M-estimation and solves it with an iteratively reweighted least squares (IRLS) procedure. To downplay the role of outliers Ranfl and Koltun \cite{Ranftl_2018_ECCV} proposed to cast the robust function learnable and simulated the IRLS procedure in their Deep Fundamental Matrix Estimation (DFE) approach. In attempts to overcome the limitations of IRLS, Ikami \etal \cite{Ikami_2018_CVPR} developed a novel objective function and proposed the iteratively reweighted eigenvalues minimization (IREM) for its optimization.

\section{METHOD}
\label{sec:method}


The two-view geometry of an image pair can be represented either with the essential matrix $\m E$ if the camera intrinsics are known or with the fundamental matrix $\m F$ in the more general un-calibrated setup.  The minimal solver for the essential matrix is the 5-point algorithm \cite{Nister_2004_TPAMI}, while for fundamental matrix estimation there are 8- and 7-point algorithms \cite{LonguetHiggins1981ACA, Hartley_1994_TPAMI}. In our work we use the 8-point problem formulation as it is applicable to both the fundamental and essential matrix estimation tasks. 
Without limiting the generality of our approach, 
further derivations are provided for the fundamental matrix.

Fig.~\ref{fig:framework} depicts the overview of our proposed approach. Given the source image $\bm I_S$ and target image $\bm I_T$, it estimates the fundamental matrix $\m F$ representing the two-view geometry of the image pair. Our framework consists of the following modules:
\begin{itemize}
    \item \textbf{Feature extraction and matching.} Given the image pair, this module estimates the set of keypoint correspondences $\bm p$ and the corresponding additional information $\bm \tau$ describing the quality of matches.
    \item \textbf{Initial solution.} The module computes the initial solution $\m F_0$  with a weighted 8-point algorithm. The weights $\bm \gamma_0$ are estimated with a lightweight neural network $g_{init}$ that takes as input $\bm p$ and $\bm \tau$.
    \item \textbf{Outlier-robust two-view geometry refinement.} In this recurrent block, the epipolar residual $\bm r_{m-1}^{epi}$ is first computed, based on the solution $\m F_{m-1}$ from the previous iteration. Next, the residual $\bm r_{m-1}^{epi}$ along with correspondences $\bm p$ and additional information $\bm \tau$ serves as input to the refinement network $g_{ref}$ that updates confidence weights $\bm \gamma_m$ of the matches. The weights are then utilized in our outlier-robust IHLS solver to estimate the refined solution $\m F_{m}$.
\end{itemize}

In the following sections we provide the detailed description of the proposed pipeline.

\subsection{Feature extraction and matching}
Given the pair of images $(\bm I_S, \bm I_T)$, the first step in recovering the two-view geometry is to extract distinctive features from both images and subsequently match them. To demonstrate the potential of end-to-end training with our two-view geometry estimation modules, as a backbone model for this task we 
utilize a dense feature matching transformer LoFTR \cite{Sun_2021_CVPR}. It gets as input the image pair, performs feature extraction, coarse-level matching and then refines confident matches to a sub-pixel level. As a result, it outputs a set of matches $\bm p$ as well as 
 their confidence levels $\bm w \in \br{0, 1}$. 

For the set of matched keypoints $\bm p$, we compute the following quantities as the additional information $\bm \tau \in \mathbb{R}^{N \times 4}$:
\begin{enumerate}
    \item coarse-level matching confidences $\bm w$
    \item $L_2$ coarse-level descriptor distances $\bm d_2$
    \item $cosine$ coarse-level descriptor distances $\bm d_{cos}$
    \item fine-level standard deviation of match position in the second image $\bm \sigma$
\end{enumerate}

It is worth noting that our robust fitting method does not depend on the specific type of feature extraction or matching, and can be successfully combined with both classic and learnable pipelines.

\subsection{Initial solution}


If a 3D point is observed in a pair of images $\bm I_S, \bm I_T$, where it has the homogeneous coordinates $\mathbf{x} = (x, y, 1)$ and $\mathbf{x}' = (x', y', 1)$, then such coordinates satisfy the so called epipolar constraint \cite{hartley_zisserman_2004}:
\begin{equation}
\mathbf{x}'^T {\m F} \mathbf{x} = (\mathbf{x}'^T \otimes \mathbf{x}^T) \vc{\m F} =0.
\end{equation}
In this equation ${\bf F} \in \mathbb{F} \equiv \{ \m F \in \mathbb{R}^{3 \times 3} \mid rank(\m F)=2 \}$ represents the fundamental matrix, $\otimes$ is the Kronecker product and $\vc{\cdot}$ is the vectorization operator.

Given a set of $N$ keypoint correspondences $\bm p = \{(\mathbf{x}_n, \mathbf{x}'_n)\}_{n=1}^N$, this constraint can be expressed compactly in the form of a homogeneous linear system
\bal
\bm r \equiv \m A\bm f = 0, 
\eal
where $\bm f=\vc{\bm F}$, $\m A\in \R^{N\times 9}$ and each row
\bal
\m A_{n} = \left[ x_n'x_n, x_n'y_n, x_n', y_n'x_n, y_n'y_n, y_n', x_n, y_n, 1 \right]
\eal
corresponds to the $n$-th matching pair $(\mathbf{x}_n, \mathbf{x}'_n)$. 

Then the 8-point algorithm estimates the fundamental matrix as a solution of the following \textit{Least Squares} problem:
\begin{equation}
\label{eq:argminf}
\bm f^\star = \argmin_{\norm{\bm f}{2}=1}\norm{\m A\bm f}{2}^2 = \argmin_{\norm{\bm f}{2}=1}\sum_{n=1}^N r_n^2 = \bm u_{min}(\m A),
\end{equation}
where $r_n \equiv r(\mathbf{x}_n, \mathbf{x}'_n, \bm f) \coloneqq (\m A \bm f)_{n}$ is the residual error for the $n$-th keypoints pair and $\bm u_{min}(\m A)$ is the singular vector of $\m A$ that corresponds to its smallest singular value.

The solution of~\eqref{eq:argminf} does not necessarily lead to a fundamental matrix of rank 2, as required. Therefore, the final estimate is obtained by projecting $\bm f$ on the space of rank-2 matrices. Specifically if we express the estimated matrix using SVD as $\bm f=\m U\diag{\bm\sigma_1,\bm\sigma_2,\bm\sigma_3}\m V^\transp$, with the singular values being in decreasing order, then its projection is equal to $\m F = \Pi\pr{\bm f} \coloneqq \m U\diag{\bm\sigma_1,\bm\sigma_2,\m 0}\m V^\transp$.

We are interested in outlier-robust solution, however as noticed in \cite{Ikami_2018_CVPR} robust estimation algorithms are sensitive to the initialization. To alleviate this problem, we first estimate the initial solution $\m F_0$ and then perform the iterative refinement as described in the next section.

To estimate the initial solution we solve~\eqref{eq:argminf} with weights $\bm \gamma_0 \in \R^N$ to suppress the influence of outliers:
\begin{equation}
\label{eq:argminf0}
\bm f_0 = \argmin_{\norm{\bm f}{2}=1}\norm{\diag{\bm \gamma_0} \m A\bm f}{2}^2
\end{equation}

The weights $\bm \gamma_0$ are estimated with the PointNet-like \cite{Qi_2017_CVPR} neural network $g_{init}(\bm p, \bm \tau)$ that processes $\bm p$ and $\bm \tau$ with interleaving 1D Convolution and Context Normalization layers. The network architecture is close to \cite{Ranftl_2018_ECCV} with the exception that the last layer is the sigmoid function, instead of softmax.

\subsection{Outlier-Robust Two-View Geometry Refinement}
Minimizing a quadratic function of the residual error can over-penalize the outliers and lead to a sub-optimal solution. For this reason, alternative robust loss functions have been studied. In this setup the solution can be recovered as:
\bal
\bm f^\star = \argmin_{\norm{\bm f}{2}=1}\rho\pr{\m A\bm f;\bm \theta} = \argmin_{\norm{\bm f}{2}=1}\sum_{n=1}^N \rho\pr{r_n; \bm \theta},
\label{eq:argmin_f}
\eal
where $\rho$ is a robust loss function of the residual error $\bm r$ with parameters $\bm \theta$. 

The use of a robust function $\rho$ should lead to solutions that exhibit zero residual errors for all the inliers and non-zero residuals otherwise. Moreover, for outliers we expect the robust loss to be agnostic to the level of residual error, which is not the case if the quadratic loss is employed in~\eqref{eq:argminf}. In other words, we aim for a solution $\bm f^\star$ which leads to a vector of residual errors $\bm r$ that attain a certain level of sparsity, related to the total number of inliers. 

To accomplish this, we could resort to the $\ell_1$ penalty, which corresponds to the tightest convex relaxation of the sparsity measure~\cite{Donoho2006}, 
or to non-convex penalties that can lead to an even better approximation. Among those is the $\ell_p^p$ pseudo-norm for $p\in\pr{0, 1}$~\cite{Chartrand2007}:
\bal\label{eq:lpp_rf}
\rho\pr{\bm r; \epsilon, p} = \sum_{n=1}^N \pr{ r_n^2 + \epsilon}^{p/2},
\eal
where $\epsilon > 0$ guarantees numerical stability.

Nevertheless, in all these cases certain outliers can still be over-penalized, and thus affect the accuracy of the estimation. To downplay the role of such an outliers, we introduce a weighted extension of $\ell_p^p$, which we define as
\bal\label{eq:wlpp_rf}
\rho\pr{\tilde{\bm r}; \epsilon, p} = \rho\pr{\bm \gamma \odot \bm r; \epsilon, p} = \sum_{n=1}^N \pr{\pr{ \gamma_n  r_n}^2 + \epsilon}^{p/2}. 
\eal
Here $\bm \gamma$ is a vector of non-negative weights and by $\odot$ we denote the point-wise multiplication.


The solution of the optimization problem \eqref{eq:argmin_f} with the specified robust loss function \eqref{eq:wlpp_rf} is then estimated in our IHLS solver, the detailed explanation of which is provided in the next section:
\bal
\bm f^\star = \text{IHLS}(\bm p, \bm \gamma; \epsilon, p) = \text{IHLS}(\m A; \epsilon, p)
\eal
For simplicity, hereinafter we assume that for the observation matrix it holds $\m A = \m A(\bm p, \bm \gamma)$. Indeed, it can be shown that
\bal
\tilde{\bm r} \equiv \bm \gamma \odot \bm r = \bm \gamma \odot \pr{\m A \bm f} = \diag{\bm \gamma} \m A \bm f = \tilde{\m A} \bm f
\eal

The optimal weight $\bm \gamma$ could be directly derived from the optimal residual error $\tilde{\bm r}$. On the other hand, the optimal residual can not be attained without clear separation of the correspondences into inliers and outliers. Therefore, we propose to adjust the weights $\bm \gamma$ in a recurrent manner. 

Specifically, given the solution $\m F_{m-1}$ from the previous iteration $m - 1$, we first calculate the epipolar residual $\bm r_{m-1}^{epi}$ as the \textit{Symmetric Epipolar Distance} \cite{hartley_zisserman_2004}. The $\bm r^{epi}$ can be seen as the normalized version of $\bm r$ to preserve the constrained input range for the network. We then process $\bm r_{m-1}^{epi}$  with the network $g_{ref}(\bm r_{m-1}^{epi}, \bm p, \bm \tau)$ to obtain the updated weights $\bm \gamma_m$. The architecture of the $g_{ref}$ is similar to $g_{init}$. Note also that we employ a shared weights setup, so in each iteration we use the same network $g_{ref}$.

With the new weights $\bm \gamma_m$ we find the outlier-robust solution $\m F_m$ with our IHLS solver and proceed to the next iteration. The final solution $\m F$ and the weights $\bm \gamma$ are then simply the output of the last iteration. Based on our experiments, we observed that an initialization step and 2 iterations of the recurrent unit are enough for the convergence of our framework.
\subsection{Iterative Homogeneous Least Squares Solver}
\label{sec:IHLS}

In this section we describe in detail our approach for solving the optimization problem~\eqref{eq:argmin_f} with the proposed outlier-robust loss function defined in Eq.~\eqref{eq:wlpp_rf}. To proceed, we utilize
the following inequality~\cite{Sun2017}:
\bal
\abs{f}^p \le \frac{p}{2}\frac{f^2}{\beta^{2-p}} + \frac{2-p}{2}\beta^p,
\label{eq:lp_ineq}
\eal
which holds true $\forall f\in \R$, $\beta > 0$ and $0<p\le 2$.

Let us now consider the positive scalar $ \sqrt{f_n^2+ \epsilon} $ with $ \epsilon >0 $ being a small positive constant and plug it in~\eqref{eq:lp_ineq}. Then, we obtain
\bal
\pr{f_n^2+\epsilon}^{\frac{p}{2}} \le \frac{p}{2}\frac{f_n^2}{\beta_n^{2-p}}+\frac{p\epsilon}{2}\beta_n^{p-2}+ \frac{2-p}{2}\beta_n^p.
\label{eq:lp_gamma_ineq}
\eal
The above inequality is closed under summation and, thus, it further holds that:
\bal
\rho\pr{\bm f;\epsilon, p}=&\suml_{n=1}^N \pr{f_n^2+\epsilon}^{\frac{p}{2}}
\le \frac{p}{2}\bm f^\transp\pdiag{\bm\beta}{p-2}\bm f\nonumber\\
+&\suml_{n=1}^N\pr{\frac{p\epsilon}{2}\beta_n^{p-2}+\frac{2-p}{2}\beta_n^p}=\psi\pr{\bm\beta}.
\eal
Moreover, by noting that all the terms of $\psi\pr{\bm\beta}$ are strictly positive, we end up with the following identity 
\bal
\rho\pr{\bm f;\epsilon, p} =\min_{\bm\beta\in\R^N_+} \psi\pr{\bm\beta},
\label{eq:var_beta}
\eal
which can be easily verified by differentiating w.r.t $\bm\beta$.
We have, thus, managed to obtain a variational formulation of the robust function $\rho\pr{\bm f;\epsilon, p}$ as the minimum of a quadratic function in $\bm f$ over the auxiliary variable $\bm \beta$.

Combining~\eqref{eq:argmin_f} and \eqref{eq:var_beta} we get:
\bal
\min_{\norm{\bm f}{2}=1}&\rho\pr{\m A\bm f;\epsilon,p}=\min_{\norm{\bm f}{2}=1,\,\bm\beta\in\R^N_+} \phi\pr{\bm f, \bm\beta} \equiv\nonumber\\&\frac{p}{2}\bm f^\transp\bm \Gamma\pr{\bm\beta}\bm f+\suml_{n=1}^N\pr{\frac{p\epsilon}{2}\beta_k^{p-2}+\frac{2-p}{2}\beta_k^p},
\label{eq:min_min}
\eal
where 
\bal
\bm\Gamma\pr{\bm\beta}=\m A^\transp\pdiag{\bm\beta}{p-2}\m A = \m A^\transp\m B\m A.
\label{eq:Gamma_mat}
\eal
By exploiting the variational formulation of~\eqref{eq:var_beta} we have achieved to express our original problem in~\eqref{eq:argmin_f} as a joint minimization in the variables $\pr{\bm f, \bm \beta}$, for which it is possible to consider the following alternating minimization strategy:
\begin{subequations}
\label{eq:cyclic_min}
\bal
\begin{split}\label{eq:beta_min}
\bm \beta-\textrm{step}: \bm\beta^{k+1}&=\argmin_{\bm\beta\in\R^N_+}\phi\pr{\bm f^k,\bm\beta},
\end{split}\\
\begin{split}\label{eq:x_min}
\bm f-\textrm{step}:\bm f^{k+1}&=\argmin_{\norm{\bm f}{2}=1}\phi\pr{\bm f,\bm\beta^{k+1}}.
\end{split}
\eal
\end{subequations}

The minimizer of~\eqref{eq:beta_min} can be found by equating the gradient of the respective objective to zero, $\grad{\phi\pr{\bm f^k;\bm\beta}}{\bm\beta}=\bm 0$, which takes the form:
\bal
\kappa\pr{\pdiag{\m A\bm f^k}{2}+\epsilon\bm I - \pdiag{\bm\beta}{2}}\bm\beta^{p-3} = \bm 0,
\label{eq:beta_system}
\eal
where $\kappa=p(p-2)/2$.
For the case of interest $0 < p \le 1$ and given that $\beta_n$ is constrained to be strictly positive, the only admissible solution of Eq.~\eqref{eq:beta_system} is 
\bal
\beta_n^{k+1}=\sqrt{\pr{\bm e_n^\transp\m A\bm f^k}^2+\epsilon},\,\,\forall n=1,\ldots, N,
\label{eq:beta_k}
\eal
with $\bm e_n$ denoting the unit vector of the standard $\R^N$ basis.

Next, the minimizer of~\eqref{eq:x_min} is the solution of a unit-norm constrained homogeneous least-squares problem~\cite{Szeliski2022}, and is obtained as:
\bal
\bm f^{k+1} = \bm u_{\text{min}}\pr{\bm\Gamma\pr{\bm\beta^{k+1}}},
\label{eq:f_argmin_beta}
\eal
where $\bm u_{\text{min}}\pr{\bm\Gamma\pr{\bm\beta^{k+1}}}$ is the eigenvector associated to the smallest eigenvalue of the symmetric positive definite matrix $\bm\Gamma\pr{\bm\beta^{k+1}}$.

\begin{figure}[!t]
    \centering
    \includegraphics[width=\linewidth]{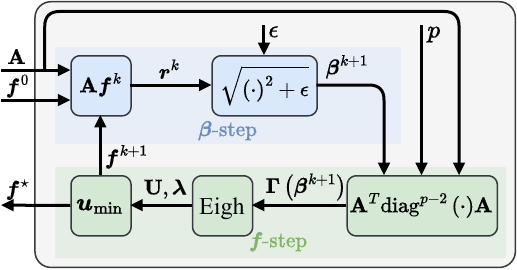}
    \caption{The proposed IHLS layer architecture. $\bm A$, $\bm f^\star$ and $\cbr{p, \epsilon}$ are the input, output and parameters of the layer, respectively, while $\bm f^0$ serves as the initialization.}
    \label{fig:ihls}
    \vspace{-0.5cm}
\end{figure}

Finally, by combining~\eqref{eq:Gamma_mat}, \eqref{eq:beta_k}, and \eqref{eq:f_argmin_beta}, the overall minimization algorithm can be written in the compact form:
\bal
\label{eq:forward_iterations}
\bm f^{k+1}=\bm u_{\text{min}}\pr{\m A^\transp \pdiag{\pr{\m A\bm f^k}^2 + \epsilon\bm I}{\frac{p-2}{2}}\m A}.
\eal
The derived update strategy corresponds to an iterative homogeneous least-squares solver (IHLS), where in each iteration $k\!+\!1$ the diagonal matrix $\m B^k=\pdiag{\pr{\m A\bm f^k}^2 + \epsilon\bm I}{\frac{p-2}{2}}$ is updated according to the solution of the previous iteration, $\bm f^k$. We provide a detailed schematic of IHLS in Fig.~\ref{fig:ihls}.

\subsection{Supervision}
\label{sec:supervision}

We supervise two-view geometry estimation with $\mathcal{L}_{\text{twg}}$ that consists of two parts, pose supervision with $\mathcal{L}_{\text{pose}}$ and direct epipolar geometry supervision with $\mathcal{L}_{\text{epi}}$:
\bal
\mathcal{L}_{\text{twg}}(\m F) = \mathcal{L}_{\text{pose}}(\hat{\m T}, \m T) + \lambda_{\text{epi}}\mathcal{L}_{\text{epi}}(\m F, \bm p_{gt}),
\label{eq:twg_loss}
\eal
where $\lambda_{\text{epi}}$ is the balancing hyper-parameter.

The pose loss $\mathcal{L}_{\text{pose}}$ is computed between the ground-truth relative pose $\bf \hat{T} = \left[ \hat{\mathbf{R}}, \hat{\mathbf{t}} \right]$ and the pose $\bf T = \left[ \mathbf{R}, \mathbf{t} \right]$ extracted from the fundamental matrix given the ground-truth camera intrinsics. It is defined as the angular distance between the translation vectors $(\hat{\mathbf{t}}, \mathbf{t})$ and the angular distance between the rotation matrices $(\hat{\mathbf{R}}, \mathbf{R})$ balanced by the factor $\lambda_{\text{rot}}$:

\bal
& \mathcal{L}_{\text{pose}}(\m F)=\cos ^{-1}\left(\frac{\hat{\mathbf{t}} \cdot \mathbf{t}}{\left\|\hat{\mathbf{t}}\right\|_2 \cdot\left\|\mathbf{t}\right\|_2}\right) \\
&\quad+\lambda_{\text {rot }} \cos ^{-1}\left(\frac{\operatorname{tr}\left(\hat{\mathbf{R}}^{\top} \mathbf{R}\right)-1}{2}\right) .
\eal

For the epipolar loss $\mathcal{L}_{\text{epi}}$ we use the \textit{Sampson Epipolar Distance} \cite{hartley_zisserman_2004} for the set of ground-truth matches $\bm p_{gt} = \{(\hat{\mathbf{x}}_v, \hat{\mathbf{x}}'_v)\}_{v=1}^V$ and the estimated fundamental matrix $\m F$:
\bal
\mathcal{L}_{\text{epi}}(\m F, \bm p_{gt}) = \frac{1}{V}\sum_{v=1}^{V}\frac{{\hat{\mathbf{x}}'^T}_v {\m F} \hat{\mathbf{x}}_v}{\norm{\m F \hat{\mathbf{x}}_v}{2}^2 + \norm{\m F^T\hat{\mathbf{x}}_v'}{2}^2}
\eal

During the end-to-end training we also utilize $\mathcal{L}_{\text{match}}$ loss for the supervision of the feature extraction and matching stages. We refer to LoFTR \cite{Sun_2021_CVPR} for the definition of $\mathcal{L}_{\text{match}}$.

The total loss $\mathcal{L}$ comprises the two-view geometry estimation loss $\mathcal{L}_{\text{twg}}$ for the matrices $\m F_m$ from each iteration of our framework plus the feature loss $\mathcal{L}_{\text{match}}$ balanced by the factor $\lambda$:
\bal
\mathcal{L}=\frac{1}{M+1}\sum_{m=0}^{M}\mathcal{L}_{\text{twg}}(\m F_m)+\lambda\mathcal{L}_{\text{match}}(\bm I_S, \bm I_T),
\eal

\subsection{Training with Implicit Backpropagation}
\label{sec:backward_pass}

To use our robust function in an end-to-end network training pipeline, it is required that gradients can be back-propagated through the IHLS layer to its inputs and parameters, which we denote altogether as $\bm \theta = \cbr{\bm A, p, \epsilon}$. Specifically, we need to compute the gradients $\nabla_{\bm \theta}L = \nabla_{\bm \theta} \bm f^\star \nabla_{\bm f^\star}L$, where $L$ is the final loss function and $\nabla_{\bm f^\star}L$ is computed according to the chain rule and its exact form depends on the operations involved in the succeeding layers of IHLS in the overall network. 

We note, that $\bm f^\star$ corresponds to the minimizer of~\eqref{eq:argmin_f} and is attained through the iterative procedure outlined in~\eqref{eq:forward_iterations}, which is  executed until convergence. A na\"ive approach would be to represent this iterative process as a recurrent computational graph, which can be unrolled using a pre-defined threshold on the total number of iterations. Backpropagation through time (BPTT)~\cite{Robinson1987} or its truncated version (TBPTT)~\cite{Kokkinos2019} could then be employed to perform the training. Such an approach can be rather inefficient since it faces two important challenges, which can limit the maximum amount of utilized unrolled iterations and compromise the convergence of the optimization process. The first one is the prohibitively large amount of memory that might be required to store the intermediate results of each one of the unrolled iterations. The second reason is related to the instabilities arising from the limited precision of numerical calculations. This results in the vanishing/exploding gradients during training of recurrent architectures~\cite{Pascanu2013}, and unstable backpropagation through eigendecomposition $\bm u_{\text{min}}$, which has to be performed for every iteration of~\eqref{eq:forward_iterations}.

To overcome all the above difficulties, we exploit the fact that any limit point of the sequence of solutions $\cbr{\bm f^k}$ produced by our proposed IHLS algorithm is a stationary point. This result is motivated by Proposition 2.7.1 in~\cite{Bertsekas1996} and the fact that IHLS is a block coordinate descent minimization method, for which a unique solution exists for both sub-problems in Eq.~\eqref{eq:cyclic_min}. Next, we utilize the Karush-Kuhn-Tucker (KKT) optimality condition~\cite{Boyd2004} associated with the original problem in Eq.~\eqref{eq:argmin_f}:
\bal\label{eq:stationarity_kkt}
\nonumber \nabla_{\bm f^\star} &\br{\mc L \pr{\bm f^\star, \lambda^\star} \equiv \rho\pr{\m A\bm f^\star;\epsilon,p} + \lambda^\star\pr{\norm{\bm f^\star}{2}^2-1}}\\
& = p \bm\Gamma\pr{\bm \beta^\star} \bm f^\star + 2 \lambda^\star \bm f^\star = \bm 0,
\eal
where $\mc L$ is the Lagrangian, $\lambda^*$ denotes the Lagrange multiplier and $\bm f^\star$ represents the stationary point produced by IHLS.
By multiplying both sides by ${\bm f^\star}^\transp$ and using that ${\bm f^\star}^\transp \bm f^\star = \norm{\bm f^\star}{2}^2 = 1$, we can compute the corresponding Lagrange multiplier $\lambda^\star$ as:
\bal
\lambda^\star = -\frac{p}{2} {\bm f^\star}^\transp \bm\Gamma\pr{\bm \beta^\star} \bm f^\star. 
\label{eq:lagrange_multiplier}
\eal
Combining Eqs.~\eqref{eq:stationarity_kkt} and \eqref{eq:lagrange_multiplier} we end up with the following necessary condition that needs to be satisfied by the output of the IHLS layer upon convergence:
\bal\label{eq:backprop_conditions}
\bm g\pr{\bm f^\star, \bm \theta} \equiv \pr{\bm I-{\bm f^\star}{\bm f^\star}^\transp}\bm\Gamma\pr{\bm \beta^\star} \bm f^\star = \bm 0,
\eal
where the dependence on $\bm \theta$ is hidden in both $\bm f^\star$ and $\bm \beta^\star$. 

Now, if we differentiate both sides of this equation w.r.t $\bm\theta$, we obtain:
\bal\label{eq:implicit_function_theorem}
&\nabla_{\bm \theta} \bm g\pr{\bm f^\star, \bm \theta} + \nabla_{\bm \theta} \bm f^\star \nabla_{\bm f^\star} \bm g\pr{\bm f^\star, \bm \theta} = \bm 0 \nonumber\\
&\Rightarrow\nabla_{\bm \theta} \bm f^\star = - \nabla_{\bm \theta} \bm g\pr{\bm f^\star, \bm \theta} \pr{\nabla_{\bm f^\star} \bm g\pr{\bm f^\star, \bm \theta}}^{-1}.
\eal
Therefore, it is possible to compute the gradient of the loss $L$ w.r.t to the set of parameters $\bm\theta$ as:
\bal\label{eq:implicit_backprop}
\nabla_{\bm \theta} L = - \nabla_{\bm \theta} \bm g\pr{\bm f^\star, \bm \theta} \pr{\nabla_{\bm f^\star} \bm g\pr{\bm f^\star, \bm \theta}}^{-1} \nabla_{\bm f^\star} L.
\eal
As a result of Eq.~\eqref{eq:implicit_backprop}, backpropagation through our IHLS layer amounts to a single vector-jacobian product (VJP) $-\nabla_{\bm \theta} \bm g\pr{\bm f^\star, \bm \theta} \bm v$, where the vector $\bm v$ is obtained as the solution of a system of linear equations $\nabla_{\bm f^\star} \bm g\pr{\bm f^\star, \bm \theta} \bm v = \nabla_{\bm f^\star} L$, and $\nabla_{\bm f^\star} L$ is the incoming gradient. We note that in both cases where our IHLS layer is applied (fundamental or essential matrix estimation problems),
 $\nabla_{\bm f^\star} \bm g\pr{\bm f^\star, \bm \theta} \in \R^{9\times9}$ corresponds to a relatively small matrix. Therefore, the computational burden introduced by the inversion of such a low-dimensional matrix is negligible. Furthermore, our derived procedure is agnostic to the number of iterations performed during the forward pass, and requires only a single backpropagation through $\bm u_{\text{min}}$ to be performed in $-\nabla_{\bm \theta} \bm g\pr{\bm f^\star, \bm \theta} \bm v$, so it addresses all drawbacks of BPTT that we highlighted earlier. We provide a detailed implementation-friendly summary of our backpropagation procedure in Algorithm 1.

We note, that in Eq.~\eqref{eq:implicit_function_theorem} of our derivations we utilize the implicit function theorem, so our backpropagation strategy belongs to the emerging class of the implicit differentiation approaches~\cite{Bai2019,Zucchet2022,Gould2022}. We find our training strategy to be similar to one proposed in~\cite{Lefkimmiatis2023}, where the implicit backpropagation rules were derived for several unconstrained minimization problems. However, this approach is not directly applicable to our case, since in Eq.~\eqref{eq:argmin_f} we are dealing with a constrained minimization problem.

\begin{figure}[!t]
    \centering
    \includegraphics[width=\linewidth]{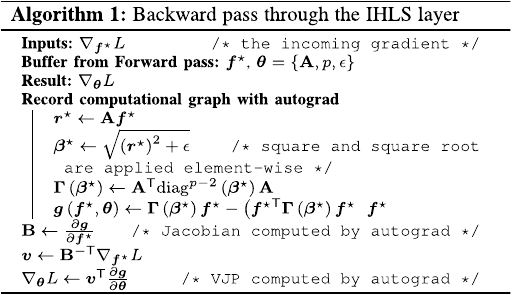}
    \label{alg:backward}
\vspace{-1.1cm}
\end{figure}
\section{EXPERIMENTS}
We evaluate the performance of the proposed approach on the outdoor and indoor relative camera pose estimation tasks. We use ScanNet~\cite{Dai_2017_scannet} for the indoor scenario and Megadepth~\cite{Li_2018_MegaDepth} for outdoor task. As the evaluation measure we use AUC of the pose error at thresholds ($5 \degree, 10 \degree, 20 \degree$) following \cite{Sarlin_2020_CVPR, Sun_2021_CVPR}. 

To perform end-to-end training of the full pipeline in the same setup that was used for training of the individual parts, we first train LoFTR \cite{Sun_2021_CVPR} from scratch following the authors' strategy. We then freeze the feature extraction and matching stages and train the two-view geometry estimation module alone. Finally, we perform end-to-end training of the feature extraction, matching and two-view geometry estimation. The training and evaluation indices both for outdoor and indoor datasets are the same as in \cite{Sun_2021_CVPR}.

All the experiments are conducted on the computational cluster of 8 NVIDIA V100 GPU's. The values of the hyper-parameters are the same for both scenarios $\lambda=1$, $\lambda_{rot}=10$ and $\lambda_{epi} = 1 \times 10^{-3}$. We use Adam optimizer with the initial learning rate of $1 \times 10^{-4}$ for the two-view geometry estimation training and $1 \times 10^{-5}$ for the end-to-end setup. 

\subsection{Baseline methods}
We perform comparison with both classic and learning-based solutions. First, we compare with RANSAC \cite{RANSAC} robust estimator, which is a standard and the most widely used algorithm for the task. We also compare with the current state-of-the-art approach MAGSAC++ \cite{Barath_2020_CVPR}. Among learning-based solutions we choose the robust loss function fitting approach DFE \cite{Ranftl_2018_ECCV} that implicitly learns an optimal robust loss function, as this method belongs to the same category as ours. Last, we compare with the recent method $\nabla$-RANSAC \cite{Wei_2023_ICCV} that improves matching confidences estimated with CLNet \cite{Zhao_2021_ICCV} by backpropagation of the learning signals through the hypotheses sampling. For the fair comparison, we trained CLNet+$\nabla$-RANSAC in the same setup as ours, with the same input $(\bm p, \bm \tau)$ to the network and the same loss \eqref{eq:twg_loss}. During inference of $\nabla$-RANSAC, the CLNet \cite{Zhao_2021_ICCV} was evaluated in combination with MAGSAC++ and Plackett-Luce model sampling following authors \cite{Wei_2023_ICCV}.

For reasons of clarity, where possible we perform hyper-parameter tuning for all the baseline methods, e.g. in MAGSAC++ we increased the default sigma threshold from 1 to 5. The learning-based methods are trained in the same setup as ours. 
\label{sec:experiments}

\subsection{Outdoor camera pose estimation}
\label{sec:outdoor}
In the outdoor scenario we use MegaDepth \cite{Li_2018_MegaDepth} dataset that consists of 1M images of 196 outdoor scenes with challenging repetitive patterns and natural illumination changes.  For training and evaluation the images are resized to the 840 pixels across the biggest dimension. The two-view geometry estimation is trained for 48 epochs with batch size 16 and end-to-end pipeline for 16 epochs with batch size 8. For the evaluation we use the same image pairs as in \cite{Sun_2021_CVPR}.

As can be seen in Table \ref{tab:outdoor}, our method outperforms classic approach MAGSAC++\cite{Barath_2020_CVPR} by a large margin (40\% at AUC@$5\degree$) and performs 3 \% better at AUC@$5\degree$ than the learning-based competitor DFE\cite{Ranftl_2018_ECCV}. Moreover, end-to-end training with matching stage further improves the results.

\begin{table}[]
\centering
\caption{Evaluation on MegaDepth \cite{Li_2018_MegaDepth} for outdoor pose estimation}
\begin{tabular}{ccccc}
\hline \hline
\multirow{2}{*}{Category}       & \multirow{2}{*}{Method} & \multicolumn{3}{c}{Pose error AUC (\%) $\uparrow$} \\ \cline{3-5} 
                                &                         & @$5\degree$          & @$10\degree$         & @$20\degree$         \\ \hline \rule{0pt}{2ex}
\multirow{2}{*}{classic}        & RANSAC \cite{RANSAC}                  & 31.53       & 46.71       & 61.38       \\
                                & MAGSAC++ \cite{Barath_2020_CVPR}               & 39.15       & 53.78       & 67.00          \\ \hline \rule{0pt}{2ex}  
\multirow{4}{*}{learning-based} & DFE (5 iterations)\cite{Ranftl_2018_ECCV}                    & 53.08       & 66.89       & 77.7        \\
                                & $\nabla$-RANSAC \cite{Wei_2023_ICCV}               & 47.83           & 61.98           & 73.34           \\
                                & Ours                    & \underline{54.86}       & \underline{68.46}       & \underline{78.94}       \\ 
                                & Ours (end2end)            & \textbf{56.63}       & \textbf{70.02}       & \textbf{80.48}       \\ \hline \hline 
\end{tabular}
\label{tab:outdoor}
\end{table}

\subsection{Indoor camera pose estimation}
\label{sec:indoor}
For the evaluation in the indoor environment we use ScanNet~\cite{Dai_2017_scannet} dataset that contains 1.5k sequences and exhibits images with low textures and challenging lighting conditions. Compared to MegaDepth\cite{Li_2018_MegaDepth}, the performance of existing approaches on ScanNet~\cite{Dai_2017_scannet} is worse as the outlier rate is higher. In our experiments the images and the corresponding depth maps are resized to the resolution of $640 \times 480$. We train two-view geometry estimation for 64 epochs with batch size 32 and end-to-end pipeline for 16 epochs with batch size 16. During evaluation we use the same image pairs as in \cite{Sun_2021_CVPR}. 

Table \ref{tab:indoor} demonstrates that our method outperforms both classic and learning-based approaches. We achieve approximately the same supremacy over the classic methods as in the outdoor scenario (37\% at AUC@$5\degree$).
And more importantly, compared to the outdoor, we achieve bigger improvement over the learning-based approaches (6\% at AUC@$5\degree$ for DFE\cite{Ranftl_2018_ECCV}). This indicates that our method handles better the cases where the outlier rate is higher.

\begin{table}[]
\centering
\caption{Evaluation on ScanNet \cite{Dai_2017_scannet} for indoor pose estimation}
\begin{tabular}{ccccc}
\hline \hline
\multirow{2}{*}{Category}       & \multirow{2}{*}{Method} & \multicolumn{3}{c}{Pose error AUC (\%) $\uparrow$} \\ \cline{3-5} 
                                &                         & @$5\degree$          & @$10\degree$         & @$20\degree$         \\ \hline \rule{0pt}{2ex} 
\multirow{2}{*}{classic}        & RANSAC \cite{RANSAC}                  & 09.76       & 22.38       & 37.04       \\
                                & MAGSAC++ \cite{Barath_2020_CVPR}                & 13.96       & 27.88       & 42.39       \\ \hline \rule{0pt}{2ex} 
\multirow{4}{*}{learning-based} & DFE (5 iterations)\cite{Ranftl_2018_ECCV}                     & 18.02       & 34.76       & 51.33       \\
                                & $\nabla$-RANSAC \cite{Wei_2023_ICCV}                & 15.24           & 29.17           & 45.38           \\
                                & Ours                    &  \underline{19.19}       & \underline{36.22}       & \underline{52.67}       \\
                                & Ours (end2end)            & \textbf{21.99}       & \textbf{39.70}       & \textbf{56.13}       \\ \hline \hline
\end{tabular}
\label{tab:indoor}
\end{table}

\section{CONCLUSIONS}
\label{sec:conclusion}
In this paper we have presented a method for outlier-robust two-view geometry estimation. Based on the feature correspondences, we first estimate the initial solution and then iteratively refine it with our outlier-robust IHLS solver. Our experiments show that the proposed approach achieves state-of-the-art results on the relative camera pose estimation task in both indoor and outdoor environments.

\section{ACKNOWLEDGEMENTS}
\label{sec:acknowledgements}
The work was supported by the Analytical center under the RF Government (subsidy agreement 000000D730321P5Q0002, Grant No. 70-2021-00145 02.11.2021)









{
    \small
    \bibliographystyle{IEEEtran}
    \bibliography{root}
}
\end{document}